\documentclass[]{style/ceurart}
\usepackage{gensymb}
\usepackage{graphicx}
\usepackage{booktabs}
\usepackage{multirow}
\usepackage{subcaption}
\usepackage{listings}
\begin{document}
\copyrightyear{2025}
\copyrightclause{Copyright for this paper by its authors.
  Use permitted under Creative Commons License Attribution 4.0
  International (CC BY 4.0).}

\conference{CLEF 2025: Conference and Labs of the Evaluation Forum, September 9-12, 2025, Madrid, Spain}

\title{Tile-Based ViT Inference with Visual-Cluster Priors for Zero-Shot Multi-Species Plant Identification}

\author[1]{Murilo Gustineli}[
orcid=0009-0003-9818-496X,
email=murilogustineli@gatech.edu,
url=https://murilogustineli.com,
]
\cormark[1]

\author[1]{Anthony Miyaguchi}[
orcid=0000-0002-9165-8718,
email=acmiyaguchi@gatech.edu,
]
\cormark[1]

\author[1]{Adrian Cheung}[
orcid=0009-0006-8650-4550,
email=acheung@gatech.edu,
]

\author[1]{Divyansh Khattak}[
email=dkhattak6@gatech.edu,
]

\address[1]{Georgia Institute of Technology, North Ave NW, Atlanta, GA 30332}
\cortext[1]{Corresponding author.}

\begin{abstract}
    We describe DS@GT’s second‑place solution to the PlantCLEF 2025 challenge on multi‑species plant identification in vegetation quadrat images.  Our pipeline combines (i) a fine‑tuned Vision Transformer ViTD2PC24All for patch‑level inference, (ii) a $4 \times 4$ tiling strategy that aligns patch size with the network’s $518 \times 518$ receptive field, and (iii) domain‑prior adaptation through PaCMAP + K‑Means visual clustering and geolocation filtering.  Tile predictions are aggregated by majority vote and re‑weighted with cluster‑specific Bayesian priors, yielding a macro‑averaged F1 of 0.348 (private leaderboard) while requiring no additional training.  All code, configuration files, and reproducibility scripts are publicly available at \href{https://github.com/dsgt-arc/plantclef-2025}{github.com/dsgt-arc/plantclef-2025}.
\end{abstract}


\maketitle

\section{Introduction}
\label{sec:intro}
The PlantCLEF task \cite{plantclef2025} within the LifeCLEF lab \cite{lifeclef2025} asks competitors to identify all plant species present in high-resolution ($\approx 2000 \text{px}$) images of $50 \times 50\text{cm}$ vegetative quadrat frames placed on the ground to determine a specific area for sampling plant species.
Recent advancements in deep learning and collaborative data platforms have created unprecedented opportunities to automate species identification across diverse ecological contexts.
The main challenges of the task lie in the domain-shift between \textbf{single-label training images} and \textbf{multi-label test images}, an extreme class imbalance with over 800 species in the test set and 7,806 in the training set, the size of the training set with 1.4 million images totaling ~281GB, and  the high intra-class variation across growth stages, organ types, and environmental contexts.
To address these challenges, we adopt a transfer-learning strategy built on the ViTD2PC24All backbone--a Vision Transformer that was first self-supervised with DINOv2 and later fine-tuned by the PlantCLEF 2025 organizers on the 2024 single-label dataset.
In this work, we investigate how far a publicly released, PlantCLEF-fine-tuned ViT can go on the 2025 multi-label task using zero-shot inference enhanced by tile-based classification, geospatial filtering, and visual-cluster Bayesian priors.

\section{Related Work}
\label{sec:related-work}

There were a total of seven teams participating in the PlantCLEF 2024 challenge \cite{plantclef2024}, in which three of those shared their solutions as working note papers.
Most participants leveraged the fine-tuned ViT provided by the organizers, as training models from scratch using the 1.4M single-label images in the training set poses significant computational challenges.
The three main methods used were: \textbf{(1)} Tiling-based inference with false positive reduction (best approach) \cite{Foy2024UtilisingDF};
\textbf{(2)} Embedding extraction and dimensionality reduction for classification \cite{gustineli2024multi}; and \textbf{(3)} Multi-label classification with composite training images \cite{Chulif2024PatchwiseIU}.

\section{Datasets and Models}
\label{sec:overview}

\subsection{Training dataset}
\begin{figure}[!htbp]
    \centering
    \includegraphics[width=0.8\textwidth]{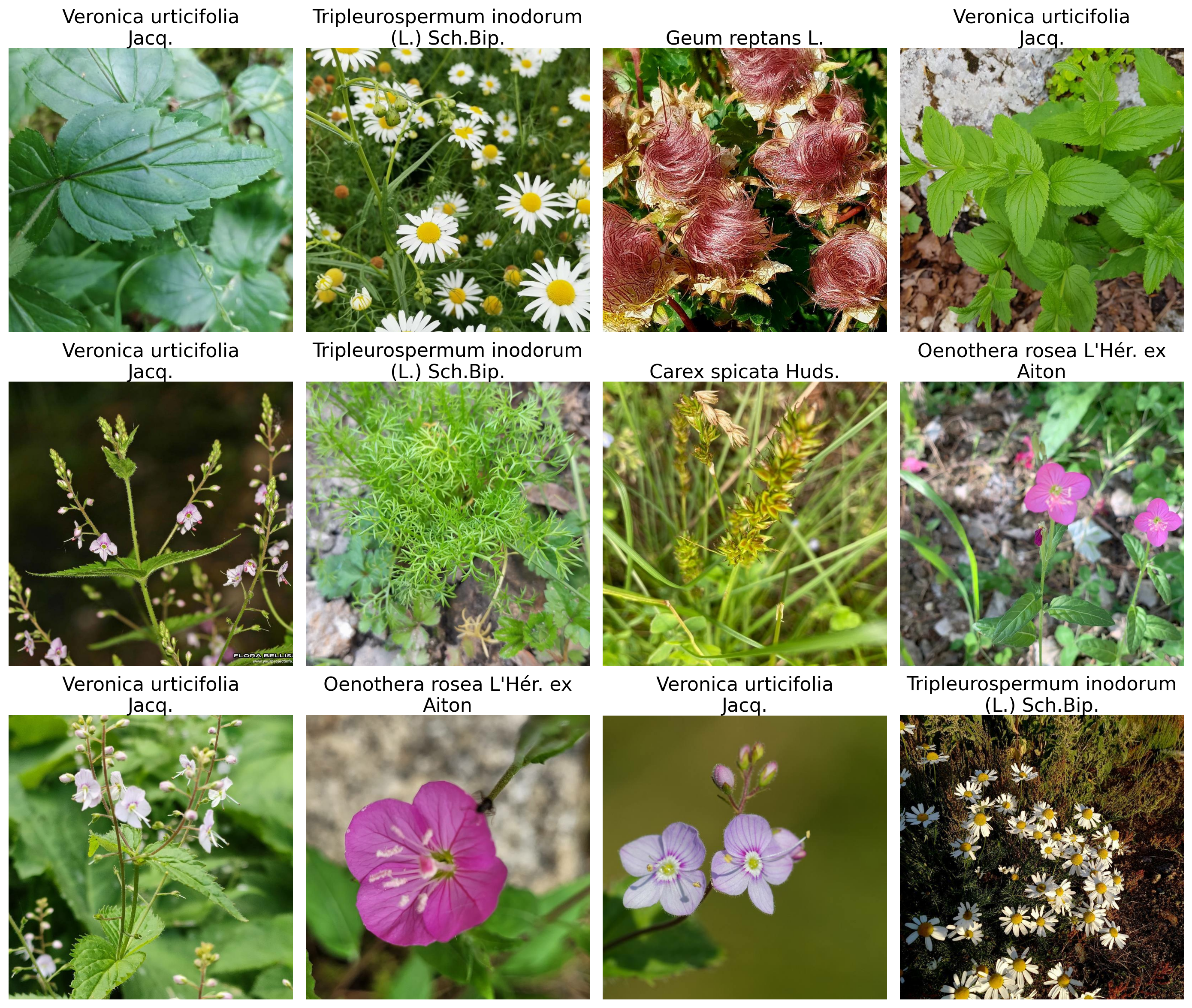}
    \caption{
    Single-label training images displaying the following six species: \textit{Veronica urticifolia Jacq., Tripleurospermum inodorum (L.) Sch.Bip., Geum reptans L., Carex spicata Huds., Oenothera rosea L'Hér. ex Aiton, Lamium bifidum Cirillo}.
    }
    \label{fig:single-label}
\end{figure}

The training dataset is a subset of the Pl@ntNet \cite{goeau2013pl} training data, composed of single-label plant species, focusing on southwestern Europe (Figure \ref{fig:single-label}).
As supplied by the organizers, the dataset comprises 7,806 plant species in 1.4 million images, totaling 281GB (Table \ref{tab:training-data}).
The high-resolution images have 800 pixels on their longest side, allowing the use of classification models that can handle large resolution inputs and facilitating the prediction of small plants in large vegetative plots.
The images were organized into subfolders by class (i.e., species) and split into predefined train/validation/test sets to facilitate the training of classification models.

\begin{table}[h!]
    \centering
    \caption{
    PlantCLEF 20204 \cite{plantclef2024} dataset overview. The dataset was split into \textbf{train/validation/test} sets to facilitate the training of classification models on the individual plant species. Note that the single-label test set used for model training differs from the challenge test set, which contains large multi-label images.
    }
    \begin{tabular}{p{1.1cm} p{1.3cm} p{1.8cm} p{0.9cm} p{1cm}} \toprule
    \textbf{Datasets} & \textbf{Images} & \textbf{Observations} & \textbf{Species} & \textbf{Genera}  \\ \midrule
    All & 1,408,033 & 1,151,904 & 7,806 & 1,446 \\ \midrule
    Train & 1,308,899 & 1,052,927 & 7,806 & 1,446 \\
    Val & 51,194 & 51,045 & 6,670 & 1,415 \\
    Test & 47,940 & 47,932 & 5,912 & 1,375 \\ \bottomrule
    \end{tabular}
    \label{tab:training-data}
\end{table}

\subsection{Test dataset}
\begin{figure}[!htbp]
    \centering
    \includegraphics[width=0.8\textwidth]{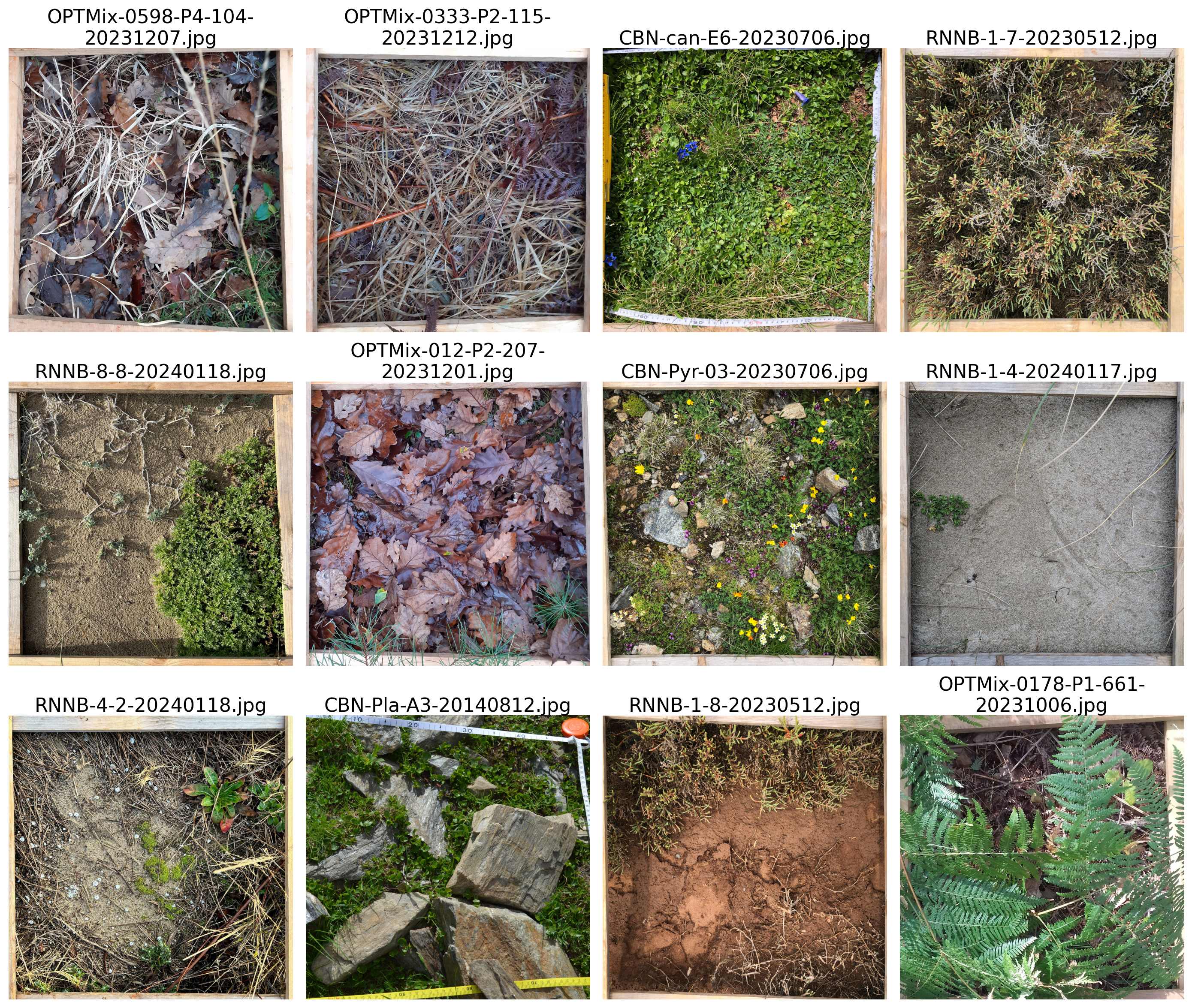}
    \caption{
    Subset of twelve test set images showcasing the significant domain shift between different quadrats.
    }
    \label{fig:test-distribution}
\end{figure}

The test set features image quadrats of many floristic environments, emphasizing Pyrenean and Mediterranean flora.
All datasets are curated by experts and include a total of 2,105 high-resolution quadrat images (Figure \ref{fig:test-distribution}).
The shooting protocols can differ considerably depending on the context, with variations such as using wooden frames or measuring tape to outline the plot, or capturing images from angles that may not be perfectly perpendicular to the ground due to the site's slope.
Furthermore, image quality can fluctuate based on weather conditions, leading to factors like pronounced shadows, blurred areas, and other visual inconsistencies.

\subsection{Fine-tuned models}
The ViTD2 and ViTD2PC24 models are Vision Transformers (ViTs) pretrained using the DINOv2 Self-Supervised Learning (SSL) approach on the LVD-142M dataset, which contains 142 million images \cite{oquab2023dinov2}.
These models were fine-tuned on the PlantCLEF 2024 dataset to address plant species identification \cite{plantclef2024}.
The original ViTD2 model serves as the backbone, pretrained with DINOv2 without the classifier head, and was not further fine-tuned on PlantCLEF data.
This model is mainly used for extracting general image embeddings.
The ViTD2PC24 models, however, build on top of the backbone with additional supervised training tailored for plant classification.

To simplify their naming, ViTD2PC24OC refers to the version where only the classifier head was fine-tuned, while ViTD2PC24All refers to the model where both the backbone and classifier head were fine-tuned.
The models were made available to participants to facilitate their experiments, particularly those with limited computational resources, and played a key role in developing solutions for the PlantCLEF 2024 challenge.
We exclusively utilized the \textbf{ViTD2PC24All} model, as it was more effective in extracting richer embedding representations and achieving higher classification scores as compared to its counterpart.

\subsection{Evaluation metric}
The task is evaluated using the \textit{Macro-Averaged F1 score per sample}, which provides a balance between precision and recall for multi-label classification.
We reproduce the formula for completeness (Eq. \ref{eq:f1}--\ref{eq:final}).
The goal is to predict the presence of one or more plant species in high-resolution quadrat images.
This evaluation metric takes the average of the F1 scores computed individually for each vegetation plot.
The F1 score for each quadrat image is calculated as the harmonic mean of precision and recall:
\begin{equation}
  \text{F1}_{j}=\frac{2 \cdot \text {Precision}_{j} \cdot \text {Recall}_{j}}{\text {Precision}_{j} + \text {Recall}_{j}}
  \label{eq:f1}
\end{equation}

Where $\text{Precision}_{j} =\frac{TP}{TP_{j} +FP_{j}}$ and $\text{Recall}_{j} =\frac{TP_{j}}{TP_{j} +FN_{j}}$
with $TP_j$, $FP_j$, and $FN_j$ denoting true positive, false positive, and false negatives for image $j$. To ensure fairness across ecological regions (transects), macro-averaging is applied in a two-step process:

\begin{enumerate}
    \item $\text{F}1$ scores are averaged across all quadrat images within each transect.
    \item These per-transect averages are then averaged across all transects to yield the final score:
    \begin{equation}
        \text{Final Score} =\frac{1}{N}\sum _{i=1}^{N}\left(\frac{1}{T_{i}}\sum _{j=1}^{T_{i}} F1^{j}\right)
        \label{eq:final}
    \end{equation}
\end{enumerate}

Where $N$ is the number of transects, ${T}_{i}$ is the number of quadrats in transect $i$, and ${F1}^{ j}$ is the $\text{F}1$ score of image $j$.


\section{Methodology}
\label{sec:methodology}

\begin{figure*}[!ht]
    \centering
    \includegraphics[width=\textwidth]{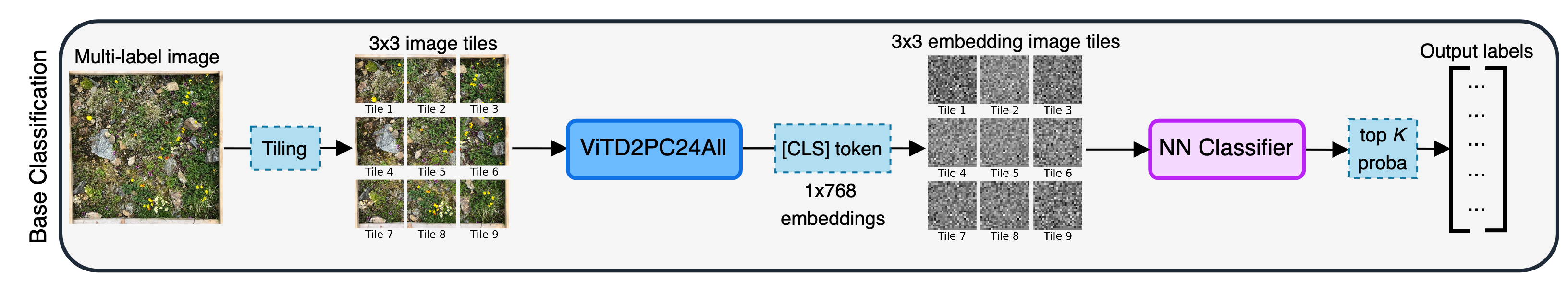}
    \caption{
    Overview of our proposed transfer learning method.
    We perform a tiling approach on the test set, classify each tile using the ViTD2PC24All model, and aggregate the results by selecting the top-K species based on their frequency count.
    }
    \label{fig:model-diagram}
\end{figure*}

Our approach leverages the embedding space learned by the ViTD2PC24All model as a generalized feature representation of images, which is used for classification (Figure \ref{fig:model-diagram}).
ViTD2PC24All learns robust feature representations by processing images as sequences of fixed-size patch tokens with an additional [CLS] token for classification tasks \cite{wu2023cls}.
These tokens serve as low-dimensional representations of the image patches, similar to words in a phrase for language models.
The main challenge lies in overcoming the domain shift between single-label training images and multi-label test images.
We perform a tiling approach, dividing each test image into a grid of $N\times N$ tiles. Our code is available at \href{https://github.com/dsgt-arc/plantclef-2025}{github.com/dsgt-arc/plantclef-2025}.

\subsection{Tiled inference}
We use the fine-tuned ViTD2PC24All model as a baseline classifier.
To bridge the gap between single-label training images and multi-label test images, we perform a tiling-based classification strategy on the multi-label test images.
We begin by leveraging the fine-tuned classification head at inference time, where each high-resolution test image is partitioned into a fixed-size grid of non-overlapping tiles (e.g., $3\times 3$ or $4\times 4$). 
Each tile is independently classified using the fine-tuned ViT model.
This method enables localized prediction within the image and helps with the mismatch between the global multi-label test images and the local single-label learning context of the model.
To produce image-level predictions, we aggregate tile-level predictions across the image and rank species based on their frequency of occurrence among the top-K predictions per tile.
The most frequent predicted species are selected as the final image-level labels.

We empirically determined the optimal grid size to be $4\times 4$ tiles.
That aligns with the input resolution of the fine-tuned model ViTD2PC24All, which is based on the \href{https://huggingface.co/timm/vit_large_patch14_dinov2.lvd142m}{\texttt{timm/vit\_large\_patch14\_dinov2.lvd142m}} architecture and expects images resized to $518\times 518$.
A typical high-resolution quadrat image has a width of approximately 2000 pixels, partitioning in $4\times 4$ tiles yields sub-images of roughly 500 pixels per side, closely matching the model's expected input size.
Using smaller or larger grid sizes leads to image downscaling or upscaling during preprocessing, resulting in degradation of feature quality and decreasing classification performance.

\subsection{Geolocation filtering}
\begin{figure}[!ht]
    \centering
    \includegraphics[width=\textwidth]{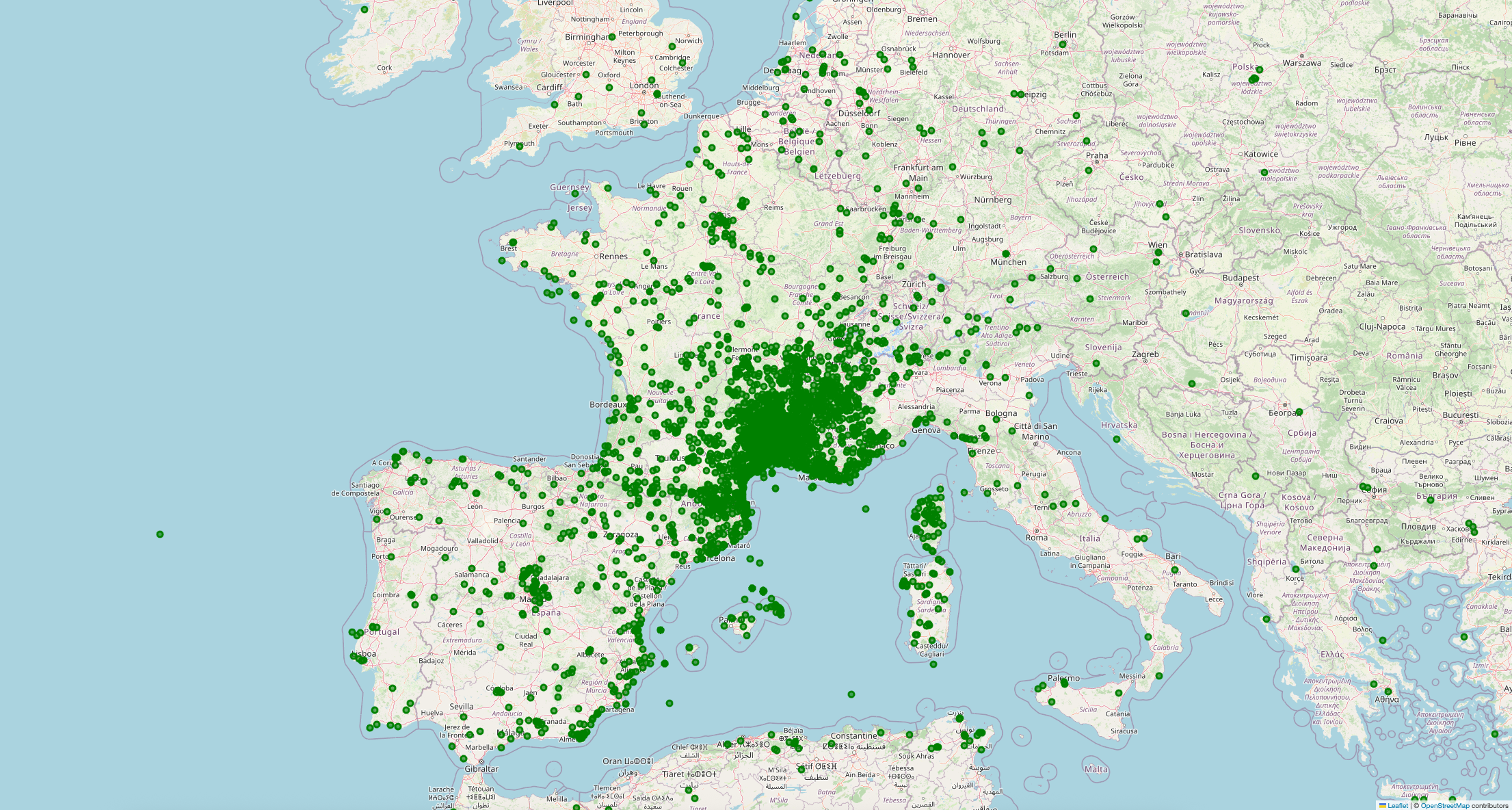}
    \caption{
    Geolocation of plant species based on their latitude and longitude metadata. 
    Using a reference point (latitude=44, longitude=4) located in the Southwestern European region, we computed the squared Euclidean distance between each plant species geolocation and the reference point, rank observations per species by proximity to the reference point, and selected the closest point ($\text{rank} = 1$) for each species.
    }
    \label{fig:geolocation}
\end{figure}

To address the domain-shift problem between training and test data -- where the training data has 7,806 species and the test data has roughly 800 species from Southwestern Europe -- we used geolocation metadata from the training images to narrow down likely species candidates.
We defined a reference point in Southern France ($44\degree \text{N}$, $4\degree \text{E}$) and computed the squared Euclidean distance between this point and each species observation. 
For each species, we selected the closest known geotagged observation.
We then filtered species whose nearest observation falls within the geographic boundaries of countries relevant to the test set (France, Spain, Italy, and Switzerland), as shown in Figure \ref{fig:geolocation}.
This geospatial filtering reduced the search space from thousands of global species to a plausible subset of 4,981 species, improving prediction relevance and mitigating the long-tailed class distribution (Table \ref{tab:ablation}).

\subsection{Visual-Cluster Bayesian prior adaptation}
To address the domain shift and class imbalance between training and test sets, we introduce a strategy to \textbf{prioritize likely species} in the test set.
We grouped test images by their corresponding region identifiers, which are present in the \texttt{quadrat\_id} field.
These identifiers represent the origin of the vegetation plots and were used to cluster images based on their location.
We defined 13 regions based on the test set \texttt{quadrat\_id} naming format and assigned each image to its respective region (Table \ref{tab:regions}).

\begin{table}[!h]
    \centering
    \caption{Grouping of the 13 regions based on the test set \texttt{quadrat\_id} naming patterns, including the number of images per region and their corresponding dominant K-Means cluster derived from visual embedding similarity.}
    \begin{tabular}{p{3.2cm} >{\centering\arraybackslash}m{2.5cm} >{\centering\arraybackslash}m{2.5cm}} \toprule
    \textbf{Region} & \textbf{Image Count} & \textbf{K-Means Cluster} \\ \midrule
        CBN-PdlC	& 816 & 2 \\
        CBN-Pla	& 628 & 3 \\
        GUARDEN-CBNMed	& 165 & 1 \\
        RNNB	& 141 & 1 \\
        LISAH-BOU	& 82 & 1 \\
        OPTMix	& 78 & 1 \\
        LISAH-BVD	& 76 & 1 \\
        GUARDEN-AMB	& 36 & 1 \\
        LISAH-PEC	& 35 & 1 \\
        CBN-can	& 30 & 2 \\
        LISAH-JAS	& 15 & 1 \\
        CBN-Pyr	& 2 & 1 \\
        2024-CEV3	& 1 & 1 \\ \bottomrule
    \end{tabular}
    \label{tab:regions}
\end{table}

\begin{figure*}[!ht]
    \centering
    \begin{subfigure}[t]{0.495\textwidth}
        \centering
        \includegraphics[width=\textwidth]{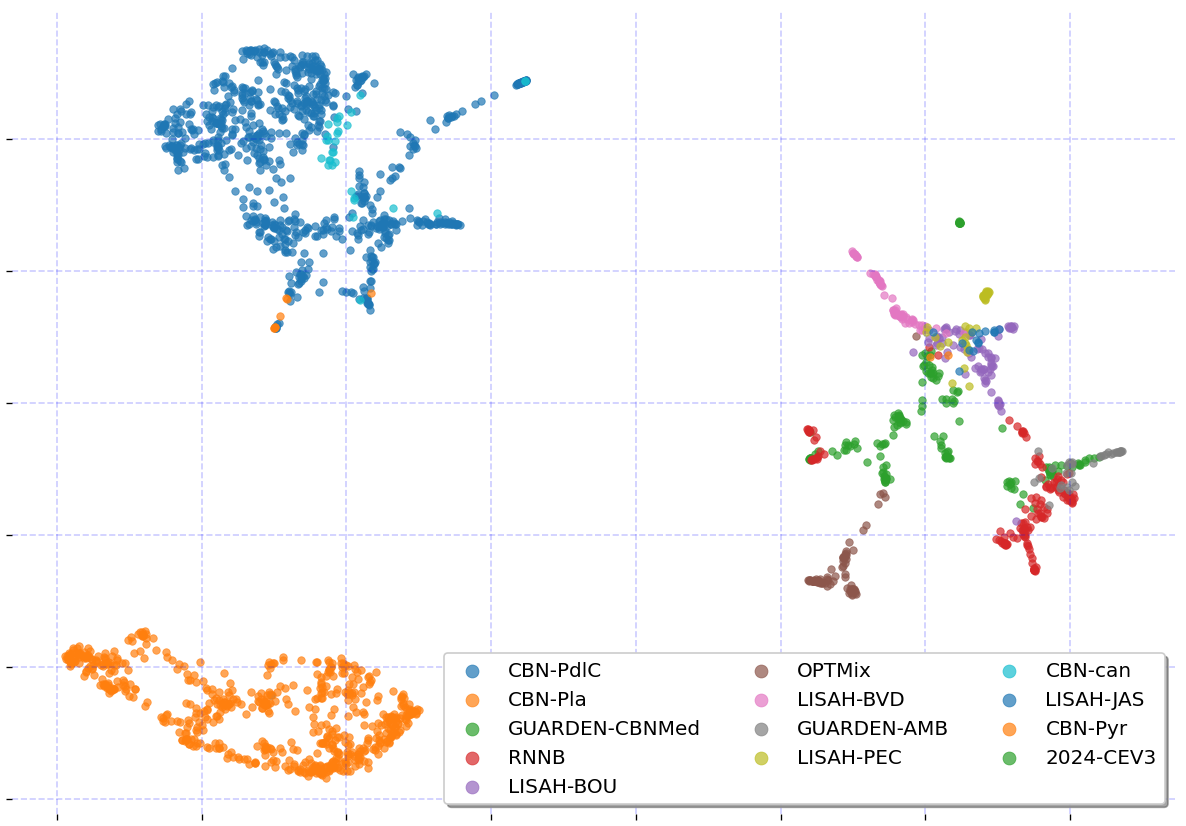}
        \caption{PaCMAP projection colored by region.}
        \label{fig:pacmap-subfig1}
    \end{subfigure}
    \hfill
    \begin{subfigure}[t]{0.495\textwidth}
        \centering
        \includegraphics[width=\textwidth]{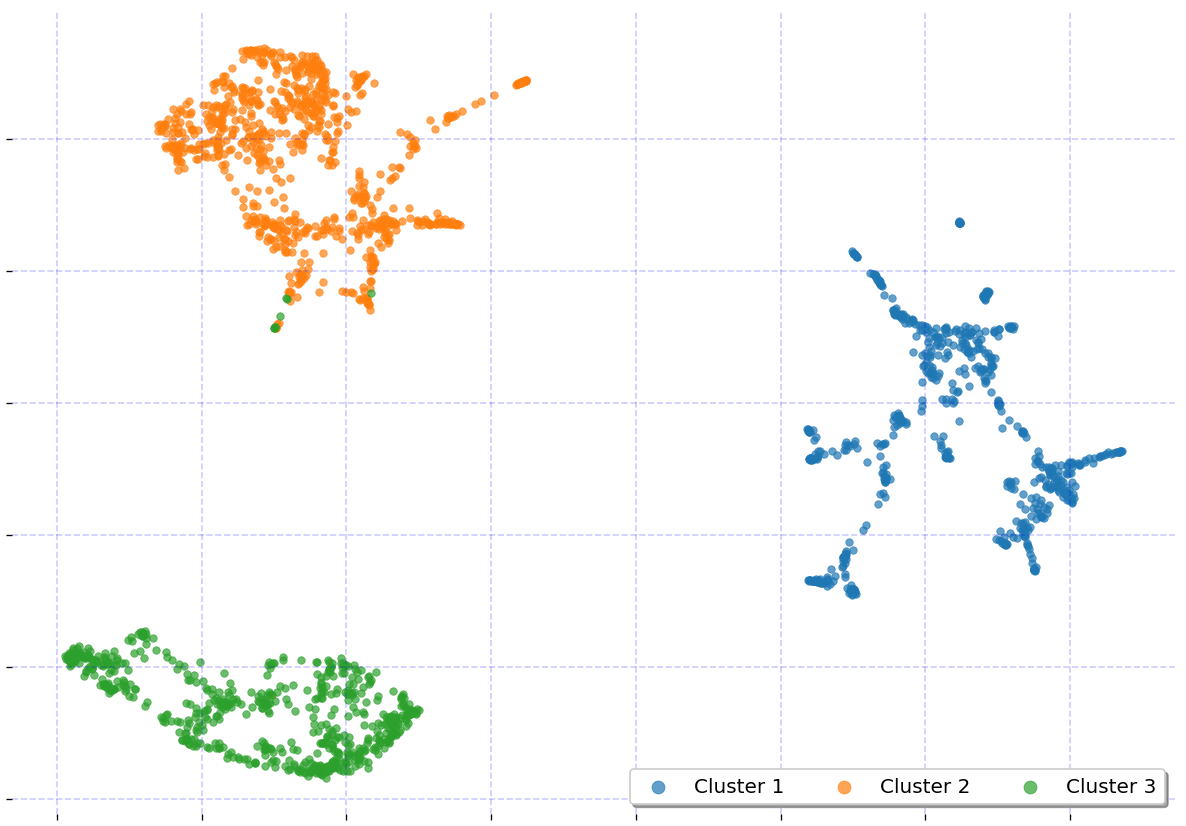}
        \caption{PaCMAP projection colored by K-Means cluster.}
        \label{fig:pacmap-subfig2}
    \end{subfigure}
    \caption{
        PaCMAP projections of the test set [CLS] token embeddings. \ref{fig:pacmap-subfig1} Two-dimensional projection of test image embeddings using PaCMAP, with each point colored by its region label (13 regions total).
        The plot reveals that quadrats from the same region tend to cluster together, indicating strong visual similarity and potential ecological coherence within regions.
        \ref{fig:pacmap-subfig2} PaCMAP projection colored by the dominant K-Means cluster assigned to each region. 
        After clustering the embeddings into three groups, we assigned each region to the cluster where most of its quadrats reside.
        This unsupervised stratification highlights coherent visual groupings that align with ecological or geographic distinctions.
    }
    \label{fig:pacmap-cluster}
\end{figure*}

We utilized the ViTD2PC24All model to extract the [CLS] token embeddings of the test set images and projected the embeddings into two dimensions using PaCMAP \cite{pacmap-JMLR:v22:20-1061} to visually explore their structure. (Figure \ref{fig:pacmap-subfig1}).
By coloring points based on their region labels, we observed that quadrats from the same region tend to cluster together, revealing three well-defined clusters.
This suggests that certain geographic or ecological similarities—such as altitude or vegetation type—may drive visual similarity among regions, which can be leveraged to improve classification performance under domain shift.

After visualizing the PaCMAP embeddings, we applied K-Means clustering to group the quadrats into three unsupervised clusters based on visual similarity.
We then assigned each region to its dominant cluster by identifying where the majority of its images were grouped (Figure \ref{fig:pacmap-subfig2}).
This provided a meaningful stratification of the test set, allowing us to model regional variation in species composition and incorporate cluster-specific priors for downstream classification.
The final region distribution in the test set is summarized in Table \ref{tab:regions}.

We hypothesize that the three dominant K-Means clusters represent different altitude levels where the test set images were taken:

\begin{itemize}
    \item \textbf{Cluster 1: “Coastal and Salt-Tolerant Plants”} -- Salt-tolerant and drought-resistant, coastal dunes, salt marshes, and sandy habitats.
    \item \textbf{Custer 2: “Alpine and Sub-alpine Specialists”} -- Hardy, low-growing plants adapted to cold, high-altitude environments (alpine meadows and rocky slopes).
    \item \textbf{Cluster 3: “Alpine Grasses and Ferns”} -- Resilient grasses and ferns, this cluster thrives in alpine grasslands and sub-alpine zones, often in rocky or well-drained soils.
\end{itemize}

To assign the descriptive labels in the bullet list, we first identified the most frequent species within each visual cluster by averaging the per-image class-probability vectors.
The top species for every cluster were then given to ChatGPT, which returned concise ecological summaries that we adopted as cluster names.

We subsequently incorporated region-specific Bayesian priors into the tile-based inference pipeline.
The PaCMAP + K-Means step yields, for every cluster $c$, an empirical prior distribution $P(y|c)$ obtained by averaging the model's predicted probability vectors across all images in that cluster.
During inference, we re-weight each tile's class probabilities by this prior, increasing the bias towards species that are visually and geographically likely for that cluster.
This approach helped narrow down the candidate species space for each test image and improve robustness to underrepresented classes.
This is particularly important given the shift from single-label training images to multi-label plot images in the test set.

\section{Results}
\label{sec:results}
We evaluated our approaches on the hidden test set provided on the Kaggle competition leaderboard.
Table \ref{tab:ablation} presents an ablation study comparing different variants of our classification pipeline.
The naive baseline--selecting the top-K most frequent species in the training data--achieved negligible performance across both public and private leaderboard splits. 
Introducing the fine-tuned ViT-based classifier without tiling improved results only marginally, highlighting the difficulty of processing high-resolution vegetation plots holistically.
Tiling test images into a $4\times4$ grid (matching the input resolution of the fine-tuned ViTD2PC24All model) led to a substantial performance gain.
Specifically, selecting the top-9 predictions per tile yielded a private leaderboard F1 score of 0.3442, representing a strong baseline for multi-label classification using patch-level aggregation.

\begin{table}[!ht]
    \caption{Ablation study of our different approaches. $4\times4$ is the tiling size.}
    \centering
    \begin{tabular}{lcccc}
        \toprule
        \textbf{Method} & \textbf{Top-K Predictions} & \textbf{Tiles} & \textbf{Private (\%)} & \textbf{Public (\%)} \\
        \midrule
        Naive baseline & top-5 & - & 0.00422 & 0.00736 \\
        Naive baseline & top-10 & - & 0.00776 & 0.00466 \\
        Naive baseline & top-25 & - & 0.00571 & 0.00440 \\
        \midrule
        ViT & top-20 & - & 0.00633 & 0.01157 \\
        ViT & top-20 & 4x4 & 0.26313 & 0.25239 \\
        ViT & top-12 & 4x4 & 0.32667 & 0.30203 \\
        ViT & top-10 & 4x4 & 0.33926 & 0.30906 \\
        ViT & top-9 & 4x4 & 0.34420 & 0.30810 \\
        \textbf{ViT + GEO} & \textbf{top-10} & \textbf{4x4} & \textbf{0.34489} & \textbf{0.31600} \\
        \textbf{ViT + PRIORS} & \textbf{top-9} & \textbf{4x4} & \textbf{0.34834} & \textbf{0.29293} \\
        \bottomrule
    \end{tabular}
    \label{tab:ablation}
\end{table}

To further mitigate the domain shift and long-tailed class distribution challenges, we incorporated two complementary strategies: \textbf{(1)} cluster-based priors derived from PaCMAP+K-Means embeddings, and \textbf{(2)} spatial filtering using geolocation priors.
Applying cluster-specific Bayesian reweighting improved the private leaderboard score to 0.3483.
Alternatively, geolocation-based filtering—removing species unlikely to occur near the test region—resulted in a private score of 0.3449 and the highest public leaderboard score of 0.3160.
These findings demonstrate that both spatially-aware inference and prior reweighting provide valuable regularization, yielding competitive performance without modifying the underlying model architecture.

\section{Discussion}
\label{sec:discussion}





Our ablation study shows that simply forwarding the full-resolution quadrat image through the fine-tuned ViT barely surpasses a frequency-based baseline ($F1 \approx 0.006$; Table \ref{tab:ablation}). 
Once the image is tiled into $4 \times 4$ sub-images whose side length roughly matches the $518\text{px}$ receptive field of ViTD2PC24All, macro-F1 jumps by two orders of magnitude (0.34). 
This finding echoes recent work on high-resolution ViTs, where tile- or window-based inference is consistently reported as the most reliable way to preserve fine-grained cues without exceeding GPU memory limits \cite{leroy2023win, li2024patchfusion}.

Adding visual-cluster Bayesian priors yields a further $+0.004$ improvement. By averaging the model’s own probability vectors inside PaCMAP + K-Means clusters, we obtain an empirical prior $P(y|c)$ that captures region-specific floristic bias; re-weighting tile probabilities with this prior is related to the "context-conditioned" re-ranking used in training-free zero-shot pipelines \cite{an2023perceptionclip} and to Bayesian reweighting strategies explored for low-shot recognition \cite{miao2024bayesian, ji2021reweighting}.
The alternative geolocation filter achieves the best public-leaderboard score (0.316) but only matches the prior-adapted model privately. 
This suggests that purely spatial heuristics over-prune plausible long-tail species that remain detectable when appearance cues and cluster priors are combined.

\subsection{Limitations and future work}
While our training-free pipeline demonstrates that tile-based ViT inference plus cluster-aware Bayesian priors can reach competitive accuracy, several limitations remain that shape our next research steps. 
First, the backbone we rely on is already fine-tuned on single-label PlantCLEF 2024 data, so our "zero-shot" claim holds only for the 2025 task; extending this strategy to domains that lack such a pre-fine-tuned model remains an open challenge.
Second, non-overlapping square tiles risk bisecting plants at tile boundaries; sliding-window inference \cite{dede2025deep}, learned token merging \cite{niu2025atmformer}, or adaptive receptive-field \cite{fayyaz2022adaptive} methods such as ViT-AR \cite{fan2024vitar} could recover boundary context without prohibitive compute.
Finally, a lightweight round of self-training on high-confidence tile pseudo-labels, or ensembling with CNN backbones that capture texture cues absent in ViTs \cite{hussain2025ensemble}, could raise the current 0.348 macro-F1 ceiling while keeping compute modest.

\section{Conclusion}
\label{sec:conclusion}

We presented a fully training-free pipeline that combines tile-based ViT inference, geolocation filtering, and visual-cluster Bayesian priors to tackle the PlantCLEF 2025 multi-label plant identification challenge.
Starting from a publicly released, PlantCLEF-fine-tuned ViT, our method boosts macro-F1 from 0.006 to 0.348 on the private leaderboard—good for second place—without updating a single model weight.
The study confirms three take-aways: \textbf{(1)} matching the inference tile scale to the ViT’s receptive field is critical for high-resolution plant imagery; \textbf{(2)} unsupervised visual clustering provides a cheap yet powerful prior that complements spatial heuristics; and \textbf{(3)} zero-training adaptation is competitive when domain-specific compute or labels are scarce. 
All code and artifacts are open-sourced to support follow-up research on even more challenging biodiversity datasets.

\section*{Acknowledgements}

We thank the Data Science at Georgia Tech (DS@GT) CLEF competition group for their support.
This research was supported in part through research cyberinfrastructure resources and services provided by the Partnership for an Advanced Computing Environment (PACE) at the Georgia Institute of Technology, Atlanta, Georgia, USA \cite{PACE}. 

\section*{Declaration on Generative AI}
The author(s) have not employed any Generative AI tools for writing this working note paper.
  

\bibliography{main}
\end{document}